\documentclass[pdflatex,sn-mathphys-num]{sn-jnl}

\usepackage{graphicx}%
\usepackage{multirow}%
\usepackage{amsmath,amssymb,amsfonts}%
\usepackage{amsthm}%
\usepackage{mathrsfs}%
\usepackage[title]{appendix}%
\usepackage{xcolor}%
\usepackage{textcomp}%
\usepackage{manyfoot}%
\usepackage{booktabs}%
\usepackage{algorithm}%
\usepackage{algorithmicx}%
\usepackage{algpseudocode}%
\usepackage{listings}%
\usepackage[font={small}]{caption}
\usepackage{subcaption}%
\usepackage{makecell}
\usepackage{todonotes}
\definecolor{applegreen}{rgb}{0.55, 0.71, 0.0}

\theoremstyle{thmstyleone}%
\theoremstyle{thmstyletwo}%

\theoremstyle{thmstylethree}%

\begin{document}

\title[Article Title]{Continual Deep Reinforcement Learning with Task-Agnostic Policy Distillation}


\author*[1]{\fnm{Muhammad Burhan} \sur{Hafez}}\email{burhan.hafez@soton.ac.uk}
\equalcont{These authors contributed equally to this work.}

\author[2]{\fnm{Kerim} \sur{Erekmen}}\email{kerim.erekmen@informatik.uni-hamburg.de}
\equalcont{These authors contributed equally to this work.}

\affil[1]{\orgdiv{School of Electronics and Computer Science}, \orgname{University of Southampton}, \orgaddress{\country{Southampton, UK}}}

\affil[2]{\orgdiv{Department of Informatics}, \orgname{University of Hamburg}, \orgaddress{\country{Hamburg, Germany}}}


\abstract{Central to the development of universal learning systems is the ability to solve multiple tasks without retraining from scratch when new data arrives. This is crucial because each task requires significant training time. Addressing the problem of continual learning necessitates various methods due to the complexity of the problem space. This problem space includes: (1) addressing catastrophic forgetting to retain previously learned tasks, (2) demonstrating positive forward transfer for faster learning, (3) ensuring scalability across numerous tasks, and (4) facilitating learning without requiring task labels, even in the absence of clear task boundaries. In this paper, the Task-Agnostic Policy Distillation (\textbf{TAPD}) framework is introduced. This framework alleviates problems (1)-(4) by incorporating a task-agnostic phase, where an agent explores its environment without any external goal and maximizes only its intrinsic motivation. The knowledge gained during this phase is later distilled for further exploration. Therefore, the agent acts in a self-supervised manner by systematically seeking novel states. By utilizing task-agnostic distilled knowledge, the agent can solve downstream tasks more efficiently, leading to improved sample efficiency. Our code is available at the repository: \url{https://github.com/wabbajack1/TAPD}.}

\keywords{Continual Learning, Reinforcement Learning, Self-Supervised Learning, Task-Agnostic Learning}



\maketitle

\section{Introduction}\label{sec1}
Animals possess the inherent capacity for lifelong learning, continuously acquiring, refining and passing on knowledge and skills through entangled neurocognitive mechanisms. These mechanisms are critical for both the development of sensorimotor skills and the consolidation and retrieval of long-term memory \cite{speranza2021dopamine}. In the complex and rapidly evolving field of artificial intelligence (AI), continual learning plays a critical role. This is particularly evident in deep reinforcement learning (RL), a subset of machine learning in which agents aim to maximize cumulative rewards by interacting with their environment \cite{Sutton1998}. The ability of these agents to learn, adapt, and apply knowledge from processing continuous streams of information in different tasks without corrupting previously acquired information is crucial \cite{miller1989ocular, parisi2019continual}. This enables them to become universal problem solvers. 

Implementing an exploratory strategy is critical for universal agents \cite{machado2018revisiting, parisi2019continual}, as it is foundational to task-agnostic learning (TAL), a paradigm where agents learn without a specific task in mind \cite{shwartz2023compress}. Curiosity allows agents to discover and learn from unfamiliar situations, which is essential for effectively accomplishing future tasks. This is particularly crucial in environments with sparse rewards, where acquiring a particular skill can be challenging due to a lack of feedback \cite{pathak2017curiosity, parisi2021interesting, Sutton1998, sutton_policy_1999}. In task-agnostic learning, agents do not have access to task-specific labels during training, focusing instead on learning general representations from the data that can be useful across a variety of tasks \cite{zhang2020task, parisi2021interesting}. This form of learning is often associated with unsupervised learning, allowing models to leverage abundant unlabeled data efficiently \cite{shwartz2023compress}. The learned representations can later be fine-tuned for specific tasks, a process known as transfer learning, enabling the development of models that can adapt to a multitude of tasks and achieve improved performance when compared to models trained from scratch. Curiosity-driven exploration, inherent to task-agnostic learning, is key to developing agents that not only respond to their environment but also actively seek out new knowledge and experiences. It prepares them to overcome unexpected challenges and to take advantage of opportunities, promoting flexibility and generalization in learning \cite{pathak2017curiosity}.

To realize the potential of universal agents, one of the biggest challenges is memory consumption, as the accumulation of knowledge can quickly overwhelm available resources, especially when systems are deployed on devices with limited memory. Another issue is scalability, as the complexity of the learning process increases with the addition of new tasks. Studies by \citet{kirkpatrick2017overcoming}, \citet{rusu2016progressive} and \citet{parisi2019continual} have highlighted these challenges and suggested possible solutions. \citet{rusu2016progressive} and \citet{schwarz2018progress} respectively, have proposed the development of algorithms capable of not forgetting and managing growing parameters. Despite promising progress and potential solutions, there are still unsolved challenges in the field of continual reinforcement learning. One of these challenges is learning without clear task boundaries. It is difficult to define when one task ends and another begins, resulting in the need for learning in a task-agnostic manner \cite{zeno2018task}. Achieving optimal performance for specific tasks also remains a challenge, as universal agents must continuously adapt to non-stationary data and refine their knowledge to achieve high performance in different contexts, seamlessly transferring knowledge across tasks \cite{pathak2017curiosity, zhao2022impact}. Finding innovative solutions to fully exploit the potential of continual learning in reinforcement learning is required.

In this paper, we introduce the \textit{Task-Agnostic Policy Distillation} algorithm, a novel learning algorithm with alternating self-supervised prediction, which addresses challenges associated with performance across tasks and learning without any specific reward function. Our approach introduces an additional phase, called the \textit{task-agnostic} phase, into the algorithmic structure of \citet{schwarz2018progress}. This phase complements the existing progress and compress phases. In the progress phase, specific tasks are learned and in the compress phase, the newly learned knowledge is compressed for later reuse. The task-agnostic phase is therefore crucial for building exploration policies and gaining generalization capabilities, which can be reused later in subsequent phases. Within the task-agnostic phase the agent explores its environments in a self-supervised manner, without external goals and then compresses its acquired task-agnostic knowledge. This task-agnostic compressed knowledge is then later used for further task-agnostic exploration, in a periodic manner, aiming at maximizing the pursuit of novel states without extrinsic rewards from the environment. Later when tasks become specific, the agent can leverage this task-agnostic consolidated knowledge for specific tasks. The concept is intuitively perceived as a reflection of the enjoyment derived from the actions, periodically, thereby consolidating the pleasure experienced. The agent explores and adapts to new environments and solves tasks faster, which enables faster knowledge transfer and thus increases the sample efficiency of the agent. We evaluate our algorithm against the baseline approach of \citet{schwarz2018progress} on Atari 2600 games from the arcade learning environment \cite{bellemare_arcade_2013}. To the best of our knowledge, this is the first work on task-agnostic policy distillation in continual reinforcement learning.

The primary contributions of our work are summarized as follows.
\vspace{-1mm}
\begin{enumerate}[topsep=0pt]
 \item We develop a novel task-agnostic policy distillation algorithm designed to learn exploratory behaviors without relying on task-specific rewards. This algorithm allows for the transfer of learned exploratory behaviors to a target policy, resulting in faster learning and improved performance on downstream tasks.
\item We develop a novel continual reinforcement learning framework that incorporates a task-agnostic phase along with progress and compress phases. This framework facilitates the learning of novel tasks over time while overcoming catastrophic forgetting in a scalable manner.
\item We evaluate our approach against three different continual learning methods across five reinforcement learning tasks from the Arcade Learning Environment. These experiments were performed in a continual learning setup where tasks are encountered sequentially.
\item In the interest of promoting transparency and reproducibility, we make our code available at \url{https://github.com/wabbajack1/TAPD}.
\end{enumerate}

\section{Related work}\label{sec2}

\subsection{Policy Distillation}
Policy distillation, introduced by Rusu et al. \cite{rusu2016policy}, serves as a fundamental technique for transferring knowledge from multiple task-specific expert policies into a single, generalized student policy. This approach reduces the computational burden of multi-task learning by compressing multiple models into one, which can perform well across various tasks. Following this, several studies have further explored and refined policy distillation techniques. For instance, Czarnecki et al. \cite{czarnecki2019distilling} examined the broader landscape of policy distillation methods, comparing various approaches and their theoretical underpinnings. They highlighted different formulations, such as entropy-regularized distillation, which allows for faster learning and better convergence properties in diverse situations. Another notable approach is the work by Watkins et al. \cite{watkins2021teachable}, where policy distillation was employed to incorporate external advice into the learning process. This method allowed for the integration of expert knowledge, enabling the agent to quickly adapt to new tasks and improve overall performance without extensive retraining. Sun et al. \cite{sun2019real} introduced real-time policy distillation in deep reinforcement learning, which aimed to distill policies continuously during training. This approach enhanced the adaptability of the learning agent by continuously integrating the distilled knowledge, thereby improving its performance across a range of tasks. These studies underscore the versatility and effectiveness of policy distillation in multi-task reinforcement learning, demonstrating its potential to significantly improve training efficiency and policy robustness across various tasks and environments.

\subsection{Continual Learning}
Continual/lifelong learning is the ability to continually acquire, fine-tune, and transfer new knowledge and skills over time. Continual learning agents face the problem of catastrophic forgetting when learning from changing input distributions, which causes the agent to forget old knowledge when learning new tasks. They are also expected to reuse previous knowledge to learn new tasks faster without retraining from scratch or re-accessing previously seen data. This is often referred to as the stability-plasticity dilemma, where stability is the ability to retain old knowledge and plasticity is the ability to acquire new knowledge. Continual learning models can be broadly categorized into three groups: 1) models that regulate intrinsic levels of plasticity \cite{schwarz2018progress, thorne2021elastic}, 2) models that dynamically change their architecture to suit the learning of each individual task \cite{rusu2016progressive, draelos2017neurogenesis, hafez2023continual}, and 3) models that use experience replay for long-term memory consolidation \cite{kemker2018fearnet, van2020brain} (see \cite{wang2024comprehensive} for a review). 

A key limitation of currently established approaches to continual learning is their reliance on static annotated data (e.g., images or texts). However, in more natural learning settings, continual learning models are required to learn continually from sequential data with meaningful temporal relations and with sparse teaching signals (annotations). Recently, a number of approaches have been proposed to achieve unsupervised continual learning, but they have primarily been designed and applied to incremental image classification \cite{rao2019continual, madaan2021representational}. Consequently, there is a need for novel models that support unsupervised continual learning for decision-making and reinforcement learning problems where task labels are sparse or unavailable. We propose to address this challenge via task-agnostic policy distillation with self-supervised prediction for efficient, continual reinforcement learning.

\subsection{Mitigating Catastrophic Forgetting}
 \citet{kirkpatrick2017overcoming} focused on the problem of catastrophic forgetting, where a neural network loses its ability to perform previously learned tasks when trained on new tasks. The authors introduce an algorithm called Elastic Weight Consolidation (EWC) that selectively slows down learning on weights that are important for previously learned tasks. The algorithm is inspired by synaptic consolidation observed in biological brains \cite{ziegler2015synaptic}. Specifically, the algorithm uses a quadratic penalty on the difference between the parameters for the old and new tasks. This penalty slows down the learning process for task-relevant weights that encode previously learned knowledge. This approach helps to preserve previously acquired knowledge. By using the Fisher Information Matrix, EWC effectively balances the need to learn new tasks while retaining performance on old tasks.
 
\citet{rusu2016progressive} introduced another concept for mitigating catastrophic forgetting with the concepts of Progressive Networks. These networks are designed to retain a pool of pretrained models during training and learn lateral connections from them to extract useful features for new tasks. This architecture is resistant to catastrophic forgetting and enables knowledge transfer across tasks. When a new task is introduced a separate neural network, named \textit{column}, is generated and appended into the Progressive Network. This new column establishes lateral connections with the existing columns. Each column corresponds to a previous task where useful features could be learned from these lateral connections when the new task is introduced. To circumvent the problem of catastrophic forgetting, the parameters associated with the previous tasks are frozen during learning. Therefore, a separate set of parameters is learned specifically for the new task.

To address the problem space in continual RL and unify different methods, \citet{schwarz2018progress} proposed an approach called ``Progress \& Compress". This approach combines multiple methods and leverages their complementary strengths within an algorithmic framework. The proposed framework consists of two neural networks: an active network and knowledge base network, which are trained in two distinct alternating phases between consecutive tasks, namely the progress and compress phases, respectively. During the progress phase, the active network utilizes lateral connections (\citet{rusu2016progressive}) from the knowledge base network. The knowledge base network contains distilled knowledge \cite{hinton2015distilling} from newly learned tasks while retaining knowledge of old tasks in the parameter space (\citet{kirkpatrick2017overcoming}). This regularization ensures that the learning parameters are similar to those adapted to older tasks in the parameter space, resulting in an average performance across all encountered tasks \cite{hafez2023continual}. This approach leverages information from previous tasks to facilitate positive forward transfer. The retention of old tasks is achieved through an online variant of the EWC algorithm, where \citet{schwarz2018progress} introduced the concept of gracefully forgetting old tasks to free up capacity for new tasks. It also addresses the lack of scalability the EWC has when dealing with a large number of tasks as it requires keeping a separate regularization term for every previous task. This is done by using a running sum of the Fisher information matrices representing the relative importance of weights to older tasks as the only regularizer. The details on the implementation of the ``Progress \& Compress" algorithm are given in Appendix \ref{secA1}.

The approach proposed by \citet{rusu2016progressive} shows promising results in preventing forgetting of previous tasks in RL. However, a limitation is the increasing number of parameters as the number of tasks grows, leading to increased computational complexity and memory requirements. Additionally, the methods in both \citet{rusu2016progressive} and \citet{kirkpatrick2017overcoming} require knowledge of the task label for inference, which may not always be available. This makes it difficult to adapt the architecture to task changes during inference, limiting its applicability in dynamic scenarios. On the other hand, \citet{schwarz2018progress} addressed several challenges in continual learning, including the ever-growing parameter space by design of their algorithm. They introduced the online variant of EWC to gracefully forget previous tasks and used lateral connections. Nevertheless, none of these approaches has addressed learning without clear task boundaries \cite{zeno2018task}.

Our proposed approach builds on the Progress and Compress framework \cite{schwarz2018progress} but extends it to enable continual learning in the absence of clear task boundaries. Additionally, we incorporate lateral connections as in \cite{rusu2016progressive} to reuse features learned from previous tasks. Unlike \cite{rusu2016progressive}, which requires storing a separate policy network for each task, our method maintains a single policy network, the knowledge base, and distills knowledge from new tasks into this network, facilitating scalable learning.

\subsection{Curiosity and Intrinsic Motivation}
As stated by \citet{chentanez2004intrinsically}, motivation is a critical element that comes in two forms: intrinsic and extrinsic. Extrinsic motivation is fueled by specific rewards, whereas intrinsic motivation arises from the inherent enjoyment or curiosity of the activity. Dopamine is a crucial neurotransmitter that influences both motivation and the pursuit of rewards. As suggested by \citet{speranza2021dopamine}, dopamine operates by binding to receptors in specific brain regions, such as the nucleus accumbens and the prefrontal cortex, thereby affecting motivation and pleasure. Dopamine is not only vital for extrinsically motivated behaviors aimed at obtaining specific rewards but also for intrinsically motivated behaviors focused on exploration and novelty \cite{chentanez2004intrinsically}. This mechanism encourages behaviors that are essential for survival from an evolutionary standpoint and also increases the efficiency of learning to solve new problems, i.e. knowledge generalization. Among animals, for instance, novel sensory stimuli can trigger dopamine cells in a manner similar to unexpected rewards \cite{dayan2002reward}. This activation tends to vanish as the stimuli become familiar, explaining why novelty is rewarding in itself.

Computational frameworks for intrinsic motivation are inspired by how human infants and children set their objectives and gradually develop skills, as highlighted by \citet{parisi2019continual}. Infants are skilled at playing and have an impressive ability to create new structured behaviors in unstructured environments that do not provide clear extrinsic reward signals. Current advancements in reinforcement learning incorporate elements of curiosity and intrinsic motivation, especially in situations with limited or sparse rewards. In environments with limited extrinsic rewards, an agent can depend on curiosity-driven exploration. As a result, the agent can explore policies and discover novel states more efficiently. \citet{hafez2017curiosity} delved deeper into this concept, emphasizing the role of intrinsic motivation in reinforcement learning. They introduced the ICAC (Intrinsically-motivated Continuous Actor-Critic) algorithm, a curiosity-driven RL approach that incrementally builds a network of local forward models. These models assist in computing the agent's intrinsic rewards. Additionally, the algorithm employs an Instantaneous Topological Map (ITM) to partition the sensory space, guiding the agent towards information-rich states and actions. The research indicates a significant performance gain for the intrinsically motivated agent compared to agents that are only motivated by extrinsic rewards.

\citet{pathak2017curiosity} proposed an approach to curiosity-driven exploration where curiosity is formulated as the error in an agent's ability to predict the consequences of its actions in a visual feature space, in a self-supervised manner. In this way, exploration is promoted by visiting states that are difficult to predict, similar to \citet{schmidhuber1991possibility}. The visual feature space is learned through a self-supervised inverse dynamics model. The agent is structured with two primary subsystems: a reward generator and a policy. The reward generator produces an intrinsic reward signal based on the prediction error of the agent's knowledge about its environment. The Intrinsic Curiosity Module (ICM), as the reward generator, consists of two neural network models: the inverse and the forward models. The inverse model aims to predict the agent's action based on the feature encodings of two consecutive states, while the forward model predicts the feature encoding of the next state based on the current state's feature encoding and the taken action.

Building upon the ICM approach, the Intrinsic Sound Curiosity Module (ISCM) developed by \citet{zhao2022impact} utilizes the power of sound in robotic actions. The ISCM provides feedback to the agent based on crossmodal prediction error, allowing them to develop robust representations and efficient exploration. This approach has shown scalability to high-dimensional input and leverages prior knowledge to accelerate learning of downstream tasks. \citet{zhao2022impact} emphasized the ability of the approach by \citet{pathak2017curiosity} to scale to high-dimensional input and utilize knowledge from past experiences for more efficient exploration and learning of unseen tasks, therefore making ICM as a reliable tool for exploring environments.

\section{Background}\label{sec3}

\subsection{Reinforcement Learning}
Let us consider a standard RL problem in which an agent interacts with a complete observable environment and adopts a strategy to maximize the cumulative future reward. An environment consists of a state space $S$, an action Space $A$, a reward function $r : S \times A \rightarrow R$, a dynamics model $p (s_{t+1}|s_t, a_t)$, and a discount factor $ \gamma \in [0, 1]$. Therefore, an RL problem is precisely described as a Markov Decision Process (MDP). Let $\pi : S \rightarrow P(A)$ be the policy, a mapping from states to probability distributions over actions. At each timestep $t$, the agent takes an action $a_t \sim \pi(s_t)$ and receives a reward $r_t = r(s_t, a_t)$ while the environment transitions into a new state $s_{t+1} \sim p(\cdot|s_t, a_t)$. A discounted sum of future rewards defines the return $R_t = \sum^{T-1}_{i=t} \gamma^{i-t} r(s_i, a_i)$. The goal is to maximize the expected return $J = \mathbb{E}_{s_0 \sim S_0}[R_0|s_0]$, where $S_0 \subseteq S$ is a set of initial states.

The action-value function is defined as $Q_\pi(s_t, a_t) = \mathbb{E}[R_t|s_t, a_t]$, and the optimal policy $\pi^\ast$ satisfies $Q_{\pi^{\ast}}(s, a) \geq Q_\pi (s, a), \forall (s, a) \in S \times A$. When the model is not available, the optimal Q-function is approximated by a neural network with parameters $\theta^{Q}$ and trained to minimize the loss $\mathcal{L}$ between the target value $y_t = r (s_t, a_t) + \gamma \max_a Q (s_{t+1}, a|\theta^{Q'})$ and the current Q-estimate, where $\theta^{Q'}$ are the target Q parameters and are updated slowly towards $\theta^{Q}$ \cite{mnih_playing_2013, li_deep_2018}:

\begin{equation}
\mathcal{L}(\theta) = (y_t - Q(s_t, a_t|\theta^{Q}))^2
\end{equation}
 
In RL, we can directly optimize a policy \( \pi \) that is parameterized by \( \theta \), with the aim of maximizing the expected return i.e. updating \(\theta\) in the direction of an estimate of the gradient \(\nabla \log \pi(a_t|s_t; \theta)R_t\). Of particular interest is the actor-critic methods, which learn a policy and a value function simultaneously, such as Advantage Actor Critic (A2C) \cite{mnih2016asynchronous, wu2017scalable}.

\subsection{Advantage Actor Critic (A2C)}
In the landscape of Reinforcement Learning (RL), the Advantage Actor Critic (A2C) algorithm occupies a prominent role, offering an elegant and efficient way to balance between action evaluation and policy optimization \cite{wu2017scalable}. A2C is a synchronous variant of the A3C algorithm, originally introduced in \citet{mnih2016asynchronous}. It falls under the category of policy gradient methods and employs an on-policy value function (the critic), denoted as \( V^{\pi}_{\theta}(s_t) \), as its training baseline. This critic function, although adding a bias, substantially reduces the variance of the policy gradient estimates, leading to quicker and more stable training.

The backbone of A2C lies in the integration of the advantage function and the synchronous training across multiple, disjoint agents, each operating in distinct environments to accumulate training samples. The intuition behind the use of the advantage function in policy gradient methods is that a step in the policy gradient direction increases the probability of actions that are better than average, while suppressing the likelihood of suboptimal actions \cite{schulman2015high}. This behavior is encapsulated by the advantage function $A^{\pi}(s_t, a_t) = Q^{\pi}(s_t, a_t) - V^{\pi}_{\theta}(s_t) = r_t + \gamma V^{\pi}_{\theta}(s_{t+1}) - V^{\pi}_{\theta}(s_t)$, which measures how the selected action \( a_t \) compares to the default behavior of the policy in state \( s_t \). Consequently, when updating the policy, the term \( \nabla_{\theta} \log \pi_{\theta}(a_t | s_t)A_t\) will point in the direction that enhances \(\pi_{\theta}(a_t | s_t)\). 
The stochastic policy \( \pi \) (also called the actor) is updated using stochastic gradient ascent. The gradient estimator for the policy is therefore
\begin{equation}
    \nabla_{\theta} J (\pi_{\theta}) = \mathbb{E}_{\pi_{\theta}} \left[ \sum_{t=0}^{T} \nabla_{\theta} \log \pi_{\theta} (a_t | s_t) A^{\pi}(s_t, a_t) \right]
\end{equation}

Subsequently, the policy parameters \( \theta \) are updated as $\theta_{k+1} = \theta_k + \alpha \nabla_{\theta} J (\pi_{\theta})$, where $\alpha$ is the learning rate and $k$ is the iteration number. The Advantage Actor-Critic (A2C) algorithm uses two neural networks with shared parameters, an Actor and a Critic, to perform decision-making tasks. In each episode, the agents choose actions in parallel based on the Actor's policy, execute them to obtain rewards, and then compute an advantage metric to measure how well those actions did compared to the average. Both the Actor and Critic are then updated based on this information. Since A2C samples actions according to the current policy and then updates the policy based on those sampled actions, it is considered an on-policy algorithm.

\subsection{Continual Learning}

Lifelong learning or Continual Learning refers to the ability of a system to continually acquire, fine-tune, and transfer knowledge over an extended period of time. This capability is essential for computational systems and autonomous agents that interact with changing environments, i.e. dynamic data distributions \cite{wang2023comprehensive}. However, the challenge lies in avoiding catastrophic forgetting, where the acquisition of new knowledge interferes with previously learned information, also rephrased as the stability-plasticity dilemma. This dilemma concerns the balance a system must maintain between its ability to learn new information (plasticity) and its need to retain existing knowledge (stability). Two types of plasticity are essential for a stable, continuous lifelong learning process: Hebbian plasticity for positive feedback and compensatory homeostatic plasticity for neural stability \cite{fox2017integrating, parisi2019continual}.

In formal, continual learning, as described by \citet{zeno2018task}, a continuously learning algorithm is confronted with a task sequence without the possibility of accessing data from past or future tasks. Specifically let $\mathcal{L} = \{ \mathcal{T}_1, \mathcal{T}_2, \ldots, \mathcal{T}_n \}$ be the continual learning space, where $\mathcal{L}$ represents the continual learning process, $\mathcal{T}_i$ represents the $i^{th}$ task to be learned and $n$ is the total number of tasks. The objective is to optimize the loss function \(J\) across all tasks while ensuring that the model does not forget the older tasks when learning new ones: $\min_{\theta} \sum_{i=1}^{n} J(\theta; \mathcal{T}_i)$. Subject to the constraint that the model parameters \(\theta\) are shared across all tasks and updated incrementally $\theta_{k+1} = \theta_k - \alpha \nabla J(\theta_k; \mathcal{T}_{i})$, where $k$ is the iteration step of the parameters $\theta$.

In continual learning, it is important to differentiate in which context the agent should solve problems, because there could be different variations in which the agent encounters problems and in which granularity the continual learning agent is solving a task. \citet{van2019three} suggested distinct scenarios for continual learning to standardize evaluation and enable more meaningful comparisons across different methods. In \textit{Task-Incremental Learning}, models are always informed about the task at hand, allowing for task-specific components in the model. Here the model is explicitly instructed which task to perform each time, enabling it to use specialized settings or tools optimized for each task. Whereas in \textit{Domain-Incremental Learning}, task identity is not available at test time. However, models only need to solve the current task, they are not required to infer which task it is. This scenario is relevant for situations where the structure of tasks remains the same, but the input distribution changes. In \textit{Class-Incremental Learning}, models must solve each task seen so far and infer which task they are presented with. \textit{Task-agnostic learning} expands on the scenario of Domain-Incremental Learning by eliminating the use of task IDs during the training phase (\citet{zeno2018task}). A task-agnostic agent is capable of learning in an environment without any extrinsic goal \cite{parisi2021interesting}, meaning no specific reward function is given during training. Specifically, this means that the agent can solve different tasks without knowing their boundaries or task-IDs, i.e. indicating which task they are solving.


\section{Task-Agnostic Policy Distillation}
In knowledge distillation \cite{hinton2015distilling, rusu2016policy}, the goal is to train a target network, referred to here as the knowledge base, to produce the same output distribution as the original network, referred to here as the active column. In the proposed task-agnostic phase, the active column is initially trained to maximize an intrinsic reward that encourages exploration, without relying on task-specific extrinsic rewards. Subsequently, the knowledge from the active column is transferred to the knowledge base by distilling the exploratory behavior of the trained active column network into the knowledge base network. This process is repeated iteratively as necessary. Below, we provide more details on the training environment and the variations of the task-agnostic phase, followed by a discussion on the intrinsic reward used for training.

\subsection{Meta-Environment}

\begin{figure}
    \centering
    \includegraphics[width=\linewidth]{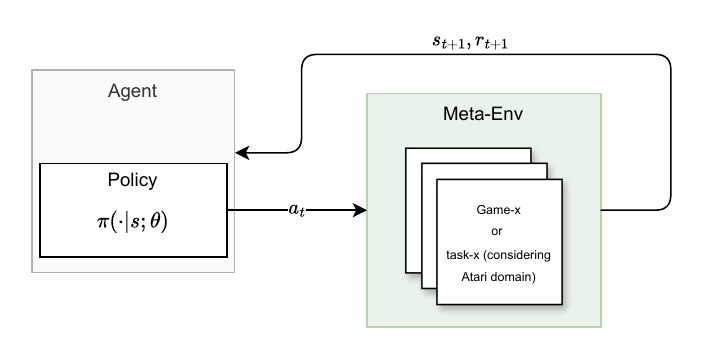}
    \caption{Illustration of the Meta-Environment, representing different game environments as tasks within the context of Atari 2600 games.}
    \label{fig:meta-env}
\end{figure}

Let $\mathcal{E}$ represent the set of all environments. For a given task $k$, let $E^k \in \mathcal{E}$ denote the environment $E$ with task $k$, where $k \in \mathbb{N}$. Within the context of continual learning, specifically in the Atari 2600 domain \cite{bellemare_arcade_2013}, it is often assumed that each environment is considered a distinct task. This concept extends to the idea of learning task $k$ within a given environment $E$ and $r_{t+1}, s_{t+1} \sim P^{E^k \in \mathcal{E}}(\cdot)$ represents the dynamics of that environment. This formalism is congruent and applicable to scenarios where different tasks exist within a single environment, such as a robot solving different tasks in an environment, where \(E^k\) signifies a task within $E$. In the case of the Atari 2600 domain \cite{bellemare_arcade_2013}, we consider all games/environments to constitute one unified environment and conceptualize $\mathcal{E}$ as a Meta-Environment, dependent upon the task context and the definition of what constitutes a task, given the versatile nature of the term task. The Meta-Environment replicates a condition where a single environment is present, encompassing diverse tasks. This concept is consistently applied throughout the paper, presenting inherent robustness to changes in environmental dynamics, such as the sudden introduction of a new game, which are interpreted as shifts within the Meta-Environment and therefore do not negatively impact the task-agnostic policy (see Section \ref{sec 4.2}). For a visual representation, see Fig. \ref{fig:meta-env}. It is important to note that the tasks can be sampled in a predefined, specified sequence from the Meta-Environment. However, this specified sequence can be relaxed to represent a specific distribution, from which the tasks (game environments) are sampled. Learning within the Meta-Environment can be more challenging due to the different textures of the tasks (i.e. game environments) and the high dissimilarity between tasks.

\subsection{Variations of the Task-Agnostic Phase}
\label{sec 4.2}
As proposed by \citet{schwarz2018progress}, the generality of the introduced progress and compress algorithm makes it a suitable candidate for integration into various frameworks. The work of \citet{schwarz2018progress} is extended to include a \textit{task-agnostic} phase. Consequently, this paper introduces the \textit{Task-Agnostic Policy Distillation} framework, which augments the progress and compress framework. Building on the foundational architecture of \citet{schwarz2018progress}, the task-agnostic phase introduces variations that enable the agent to operate without explicit task boundaries \cite{zeno2018task}. This phase implies a learning period where the agent learns without an external goal, meaning it receives no rewards from the environment, addressing the question of learning without a well-structured reward function. A curiosity-driven intrinsic reward signal is introduced through self-supervised prediction (see Section \ref{sec 4.3}), as conducted by \cite{pathak2017curiosity, zhao2022impact}. The task-agnostic phase develops its exploratory policy by maximizing the RL objective $\mathbb{E}_{\pi}\left[ \sum_{n=0}^{\infty} \gamma^n r^{i}_{t+n}\right]$, where $r^{i}_{t}$ represents an intrinsic reward generated by the agent itself at timestep $t$. This increases the probability of discovering novel states and potentially acquiring novel skills. The agent can leverage its exploratory policy and the generalized skills it induces to become a versatile problem solver when faced with specific tasks, addressing issues of positive forward transfer and enhancing sample efficiency. Specifically, the loss to be minimized in the task-agnostic phase is therefore:

\begin{equation}
\mathcal{L}^{agnostic}(\theta) = -\mathbb{E}_{\pi}\left[ \sum_{n=0}^{\infty} \gamma^n r^{i}_{t+n}\right].
\end{equation}

\begin{figure}
    \centering
    \includegraphics[width=\linewidth]{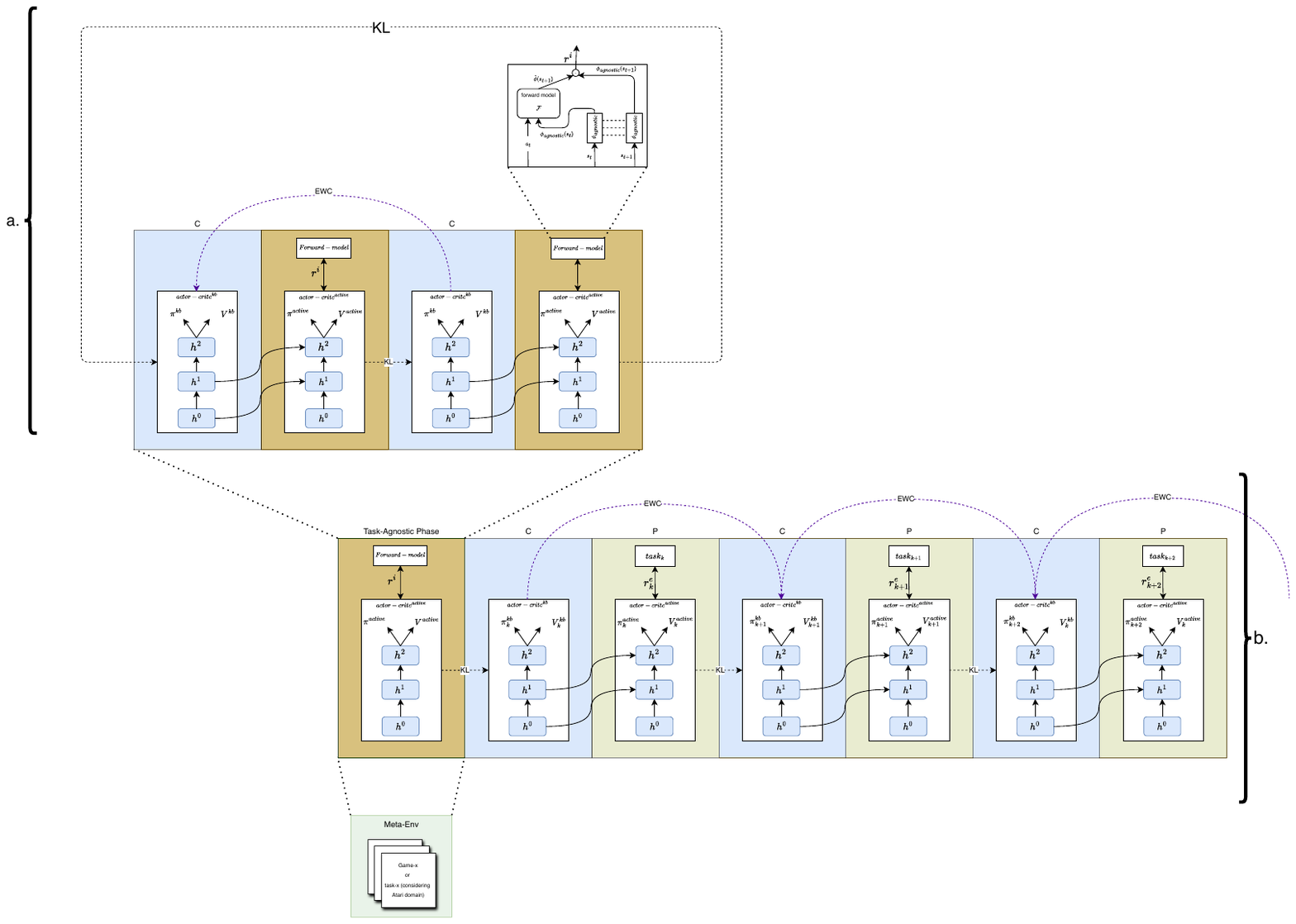}
    \caption{Overview of our Task-Agnostic Policy Distillation framework. (a) The task-agnostic phase is an abstraction of a process where intermediate alternations between maximizing intrinsic rewards and distillation occur. This process follows the same alternating pattern as in the progress and compress framework.
    (b) Here, the task-agnostic phase is initially used before alternating between progress (P) and compress (C) phases. When considering the Atari domain, each task can be randomly selected from the Meta-Environment, therefore simulating one game environment. In the C phase, the recently learned policy by the active column (green) is distilled into the knowledge base (KB) (blue) using the KL loss between the active column and KB while protecting KB's old values using Elastic Weight Consolidation (EWC). In the P phase, features learned from previous tasks are reused via lateral connections when learning new tasks. $r^e_k$ and $r^i$ are the extrinsic reward of task $k$ and the task-independent intrinsic reward, respectively. $h$ is a hidden layer.}
    \label{fig:variation1}
\end{figure}

\textbf{Variation 1 (TAPD)} Variation 1 of the task-agnostic phase is the most versatile among all the variations discussed in this work. It is suitable for diverse environments in which the action space remains consistent but the tasks vary, where the environments are conceptualized as tasks by the Meta-Environment, a collection of different game environments. In this variation, tasks (game environments) are sampled based on a specific distribution from the Meta-Environment, such as uniform sampling here. This selection mechanism implies that the agent lacks awareness of when task transitions occur, reflecting the task-agnostic scenario outlined by \citet{zeno2018task}. As depicted in Fig. \ref{fig:variation1}, this variation builds upon the framework established by \citet{schwarz2018progress}. The architectural designs within the progress and compress phases align with the baseline \cite{schwarz2018progress}. The task-agnostic phase also makes use of the active column. The first variation of the task-agnostic phase can be described as a process that leverages the A2C algorithm to minimize \(\mathcal{L}^{agnostic}(\theta)\). In this phase, the exploratory policy, i.e. the task-agnostic policy, is distilled into the knowledge base (KB) after a specified number of intermediate timesteps, followed by the selection of another game environment, highlighting the agent's unawareness of task transitions. Thus, this variation of the task-agnostic phase encompasses learning through both curiosity and distillation (see part (a) in Fig. \ref{fig:variation1}). The distilled task-agnostic policy is then used through the lateral connections to learn a new task-agnostic policy. This new policy is once again distilled and used to enhance the knowledge base with elements of curiosity. The act of distilling the exploratory distribution of one policy into another augments the exploratory behavior, as distillation has more information per training sample than hard targets \cite{hinton2015distilling}. The integration of various task-agnostic policies into the knowledge base is performed in an alternating manner, i.e.

\begin{equation}
\label{eq:alterning}
\xrightarrow[x-steps]{\text{explore/train}} \min \mathcal{L}^{agnostic}(\theta) \xrightarrow[x-steps]{\text{compress}} \theta^{kb} \xrightarrow[x-steps]{\text{explore/train}} \min \mathcal{L}^{agnostic}(\theta) \xrightarrow[x-steps]{\text{compress}} ...
\end{equation} where $\theta$ and $\theta^{kb}$ are the learning parameters of the active column and knowledge base, respectively.

This process is illustrated by the loop in Fig. \ref{fig:variation1}, within the abstraction of the task-agnostic phase (see part (a) in Fig. \ref{fig:variation1}). It is noteworthy that the Online EWC algorithm is employed throughout the process, ensuring the retention of knowledge from previous exploratory policies. Since the task-agnostic phase solely relies on the intrinsic reward \(r^i_t\) for learning, the agent does not require specific reward schemas, eliminating the need for any task-ID during the training process. This phase is suitable for sparse-reward environments as it does not require extrinsic rewards, making it an ideal pre-training step for downstream tasks. Once exploration is completed for a given number of steps, the Progress and Compress algorithm proposed by \citet{schwarz2018progress} is used for tasks that are designed to maximize extrinsic rewards (see part (b) in Fig. \ref{fig:variation1}). The progress phase uses generalized knowledge from task-independent strategies to accomplish specific tasks. An analogy for this task-agnostic phase can be made with a child exploring different rooms. After exploring a room, the child consolidates the new experiences and knowledge gained (distillation). By exploring various rooms and consolidating their understanding each time they enter a new one, the child is subsequently assigned specific tasks by their parents.

\textbf{Variation 2.} The second variation relies on configuring the general Variation 1 of the task-agnostic phase. This means configuring the Meta-Environment in such a way that it only has one task (i.e., Atari game environment) to explore. Therefore, within the task-agnostic phase, only one game environment is being distilled in alternating phases. After each intermediate timestep, the distillation process begins again, followed by exploration, as described in Sequence \ref{eq:alterning}, but without selecting different game environments after each intermediate step. This reflects the case where we only have one environment with different tasks, unlike the Atari domain where one environment has only one task, such as scoring as many points as possible.

\textbf{Variation 3.} The third variation relies on configuring Variation 2 of the task-agnostic phase. This involves setting the number of timesteps to only run for a specific number of steps and compressing the task-agnostic policy only once, before adhering to the standard protocol of the Progress and Compress method, proposed by \cite{schwarz2018progress}. This follows the steps (the downstream task loss $\mathcal{L}^{progress}(\theta)$ is given in Appendix \ref{secA1}):
\begin{equation}
\label{eq:alterning_ones}
\xrightarrow[x-steps]{\text{explore/train}} \min \mathcal{L}^{agnostic}(\theta) \xrightarrow[x-steps]{\text{compress}} \theta^{kb} \xrightarrow[y-steps]{\text{train}} \min \mathcal{L}^{progress}(\theta) \xrightarrow[y-steps]{\text{compress}}  \textbf{...}   
\end{equation}

This approach relaxes the requirement for alternating distillation of task-agnostic policies, as only one task-agnostic policy would be distilled into the knowledge base. An analogy can be drawn with an agent that explores one room, acquiring skills within it that can later be applied to different rooms.

\subsection{Intrinsic Reward} 
\label{sec 4.3}
The task-agnostic agent predominantly utilizes self-supervised prediction as its learning mechanism. By learning to predict future states from current states and actions, the agent is able to navigate and comprehend its environment, eliminating the need for extrinsic rewards. This intrinsic motivation, inspired by the ICM module of \citet{pathak2017curiosity}, is driven by the prediction error of a forward dynamics model and enables the agent to explore and acquire knowledge about its surroundings, preparing it for future tasks in diverse environments. A distinction from \citet{pathak2017curiosity} is made by excluding the inverse dynamics model of the ICM module, as this work is solely focused on learning a forward model. In alignment with \cite{pathak2017curiosity}, it is advantageous to make predictions about the next state within the feature space. Hence, let \(\phi_{agnostic}\) be the visual encoder (here, a convolutional neural network) and let \(\mathcal{F}\) be a multi-layer perceptron. If the state at timestep \(t\) is represented as \(s_t\) (raw pixels) and the action as \(a_t\), then \(\mathcal{F}(\phi_{agnostic}(s_t), a_t) = \hat{\phi}(s_{t+1})\), where $\hat{\phi}(s_{t+1})$ is the predicted feature vector of the next state. Therefore, the composition of the networks \(\mathcal{F}\) and \(\phi_{agnostic}\) forms the ICM module, excluding the inverse model. The loss used to optimize the forward dynamic model is defined as an \( L_2 \) norm of the difference between the observed and predicted feature vectors of $s_{t+1}$ i.e. \( \mathcal{L}^{forward} = \left\|\phi_{agnostic}(s_{t+1}) - \hat{\phi}(s_{t+1}) \right\|_2^2\). The overall intrinsic reward used to seek novel states is \(r^i_t = \log(\mathcal{L}^{forward} + \epsilon)\), where \(\epsilon\) is a constant added to maintain numerical stability, for values near zero.

Using the error in predicting future states as an intrinsic reward results in exploratory behavior that enables the agent to explore and acquire knowledge about its surroundings without the need for external supervisory signals, relying only on its curiosity. By encouraging visits to states that are difficult to predict (high prediction error areas), this approach fosters a directed exploration strategy aimed at improving the agent’s forward model. This strategy is particularly effective in sparse-reward environments, as it does not depend on extrinsic rewards. During the task-agnostic phase, we apply this principle to learn an exploratory policy that is periodically distilled into a knowledge base. After each distillation, a new game environment is selected where the agent further refines its task-agnostic policy using intrinsic rewards and lateral connections from the knowledge base, which now includes the recently distilled policy. This new policy is once again distilled and used to enhance the knowledge base. The distillation of the exploratory action distribution of one policy into another augments and diversifies the exploratory behavior. After the task-agnostic phase, learning downstream tasks in the subsequent progress and compress phases is accelerated due to the generalized knowledge from task-agnostic exploration stored in the knowledge base.

\section{Experimental Evaluation}
In this section, we present the experimental evaluation of our approach. First, Section 5.1 describes the Atari games that constitute the Meta-Environment to which our approach and the compared methods are applied. Next, Section 5.2 details the implementation of the task-agnostic phase and demonstrates the agent's performance and learning progress during this phase. Then, Section 5.3 compares the performance of all methods on downstream tasks in terms of learning efficiency assessed using the average game score and the entropy of the policy distribution during the progress phase, where tasks are encountered sequentially. Following this, Section 5.4 evaluates forward transfer as an indicator of adaptability by analyzing the average score on each subsequent visit to the considered tasks. Finally, Section 5.5 evaluates the computational efficiency of each method, highlighting the trade-offs between performance and resource demands.

\subsection{Environments}
As in \citet{schwarz2018progress}, the task-agnostic policy distillation framework is used in the Atari domain. In this paper, we focuses on five different Atari games: Pong, SpaceInvaders, BeamRider, DemonAttack, and AirRaid. The Meta-Environment combines these five game environments. To modify the action space, a custom action wrapper is employed. This wrapper maps the intended actions of the agent to the corresponding movements within the specific game environment. The action space is downscaled, removing unnecessary actions, as suggested by \cite{kanervisto2020action}. Table \ref{tab:action_mapping} illustrates how actions are mapped to different games/tasks. This mapping can vary depending on the specific requirements of each game/task. For instance, in the case of BeamRider, the agent selects action 2 from the policy distribution. However, the game interface executes action 3, as specified in BeamRider. As a result, the actual (game-specific) action space is hidden from the agent.

\begin{table}[ht]
    \centering
    \caption{Action mapping for different games/tasks}
    \label{tab:action_mapping}
    \begin{tabular}{lcccc}
        \toprule
        Game/Task & NOOP (0) & FIRE (1) & RIGHT (2) & LEFT (3) \\
        \midrule
        Pong & 0 & 1 & 2 & 3 \\
        BeamRider & 0 & 1 & 3 & 4 \\
        SpaceInvaders, DemonAttack, AirRaid & 0 & 1 & 2 & 3 \\
        \bottomrule
    \end{tabular}
\end{table}

These modifications to the action space prevent the attempt of ``pointless" actions. By reducing the size of the action space, the complexity of the problem is decreased, requiring fewer computational resources and potentially speeding up training time.

\textbf{Reward scaling} Due to the varying value ranges of rewards in different game environments/tasks, enhancing stability across a series of sequentially learned tasks is crucial in all environments. To address this, a single modification is made to the reward structure of the games, but only during the training phase. Given the substantial variability in the scale of scores from game to game, all positive rewards are set to be $1$, and all negative rewards to $-1$, with $0$ rewards remaining unchanged. This normalization facilitates more stable and consistent learning across varying tasks and game environments by mitigating the impact of extreme reward values, as demonstrated in \cite{mnih_playing_2013}.

\textbf{Observations} At each timestep, an agent only receives normalized visual input, which is represented as a \(84 \times 84\) grayscaled image. The environment is configured for frame skipping, allowing the agent to interact less frequently with the environment. Observations from four consecutive frames are stacked together to form an observation with enhanced temporal dimensionality. Consequently, the agent's policy receives input in the format \([N, 4, 84, 84]\), where \(N\) represents the number of environments being used in parallel for training the agent in the progress phase with the A2C algorithm. Our implementation of A2C is based on PyTorch \cite{paszke2019pytorch, pytorchrl} and incorporates concepts from \cite{stable-baselines3, avalancherl}. We used Bayesian Hyperparameter Optimization over selected configurations to maximize normalized scores across tasks. Our experiments indicate that TAPD demonstrates strong stability, with minimal performance variation across different hyperparameter settings and consistent behavior across configurations. Details on the hyperparameters of the learning architecture, algorithms, and experimental settings are provided in Appendix \ref{secA2}.

\subsection{The Intrinsic Reward in the Task-Agnostic Phase}
We evaluate the performance of the agents based on task rewards, which are specifically designed to measure active interactions. Although extrinsic rewards are recorded, they are not used during the task-agnostic training phase. Instead, the agent focuses exclusively on optimizing the intrinsic rewards it generates. The task-agnostic phase receives tasks from the Meta-Environment, which includes BeamRider and SpaceInvaders. These tasks are uniformly sampled from the Meta Environment without any task boundaries \cite{zeno2018task}. We employ TAPD's task-agnostic phase, as it is the most general of all the three variations.

\begin{figure}[!htp]
    \centering

    \begin{minipage}{\linewidth}
        \centering
        \begin{subfigure}{\linewidth}
            \centering
            \includegraphics[width=.7\linewidth]{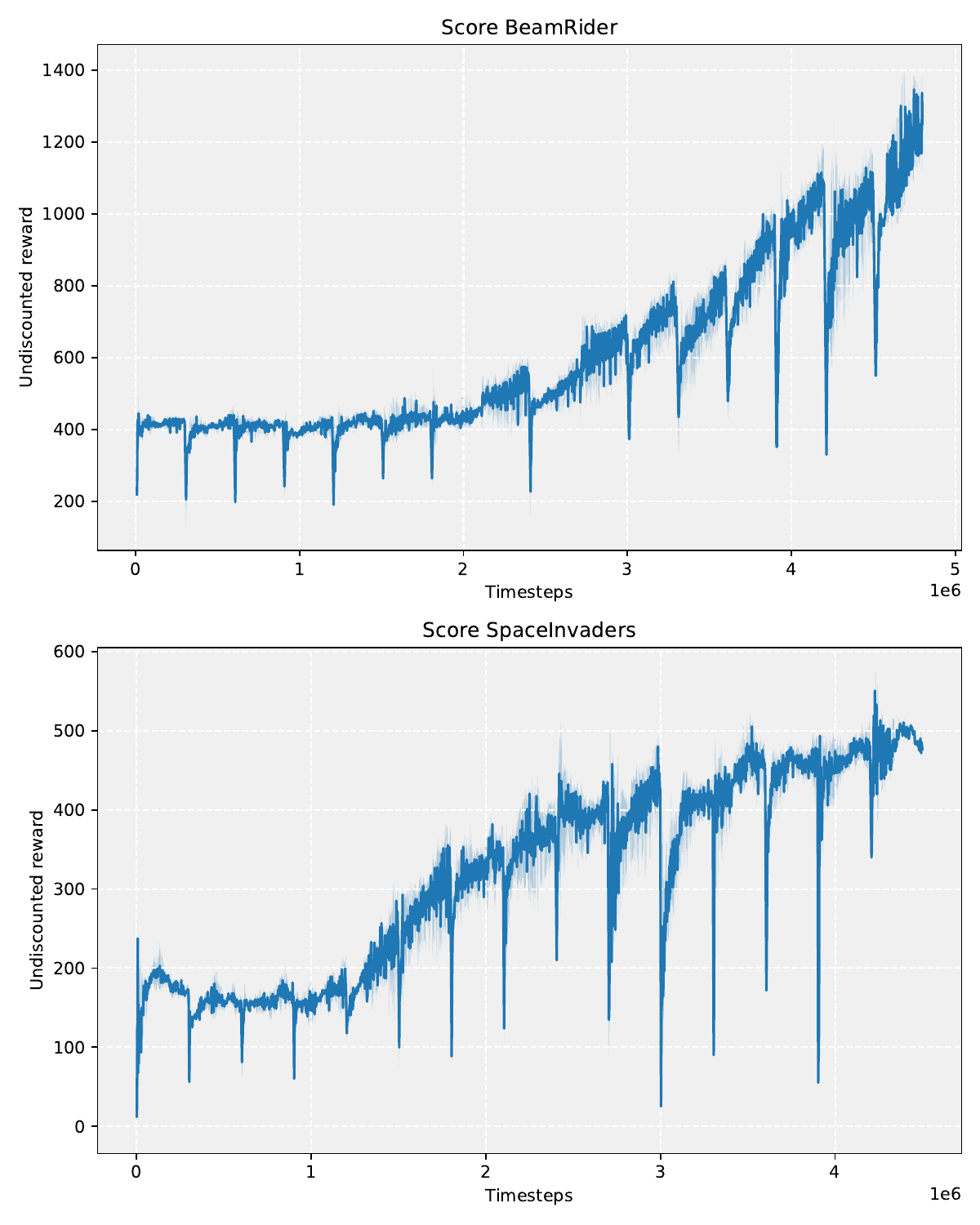}
        \end{subfigure}
    \end{minipage}
                
    \caption{Performance evaluation in the task-agnostic phase. The environment is uniformly sampled, indicating no task-boundaries. Runs averaged over 8 random seeds. Timesteps=300000 between distillation rounds in the task-agnostic phases. Averages are taken over 100 episodes.}
    \label{fig:intrinsicreward}
\end{figure}

Fig. \ref{fig:intrinsicreward} illustrates the learning progress of TAPD during the task-agnostic phase. In this phase, the task-agnostic policy is compressed after a specific number of timesteps into the knowledge base. As shown in the figure, the agent's performance improves even without extrinsic rewards. Every 300k timesteps, the task-agnostic policy is distilled into the knowledge base, resulting in an exploratory repertoire. This exploratory knowledge is then utilized within the task-agnostic phase, using generalized knowledge to further generalize knowledge. When a task change occurs, the active column in the task-agnostic phase quickly adapts to the new environment, as indicated by the spikes in the graph. This suggests that the knowledge base is aware of the environment it is in, having learned the dynamics of the environment through the forward model that produces the intrinsic rewards that guide the action selection of the active column. These rewards aid in further exploration of the environment, leading to the discovery of more novel states. The rapid task adaptation also highlights the mechanism of the knowledge base, as the active column remembers the previous situation without external goals from the environment with the help of the knowledge base. After the maximum number of timesteps in the task-agnostic phase has been reached, the alternation between progress and compress phases begins, but now with a pre-explored knowledge base, which accelerates the learning process for downstream tasks.

\subsection{Evaluation of Knowledge Accumulation and Transfer to Downstream Tasks}

We compare our Task-Agnostic Policy Distillation (TAPD) algorithm with its task-agnostic phase, Online EWC \cite{schwarz2018progress, kirkpatrick2017overcoming}, Progressive Nets \cite{xu2021adaptive}, and the Progress \& Compress baseline from \cite{schwarz2018progress}. This comparison aims to evaluate whether the task-agnostic phase can accelerate transfer learning in the progress phase for downstream tasks, while also improving sample efficiency. The task-agnostic phase, as depicted in Fig. \ref{fig:variation1}, serves as the pre-training phase. The algorithms share the same architecture and hyperparameters, with the exception that TAPD has specific parameters for the task-agnostic phase. The tasks Pong, SpaceInvaders, BeamRider, DemonAttack, and AirRaid are sequentially learned. The evaluation focuses on the learning performance in the progress phase, as the emphasis lies on evaluating the positive forward transfer and sample efficiency of the active column. In the task-agnostic phase, TAPD uniformly samples SpaceInvaders and BeamRider, simulating tasks with undefined boundaries \citet{zeno2018task}. On each visit to a task, the active columns parameters and lateral connections are re-initialized.

\begin{figure}[!htp]
    \centering
    
    \begin{minipage}{1\linewidth}
        \centering
        
        \begin{subfigure}{\linewidth}
            \centering
            \includegraphics[width=\linewidth]{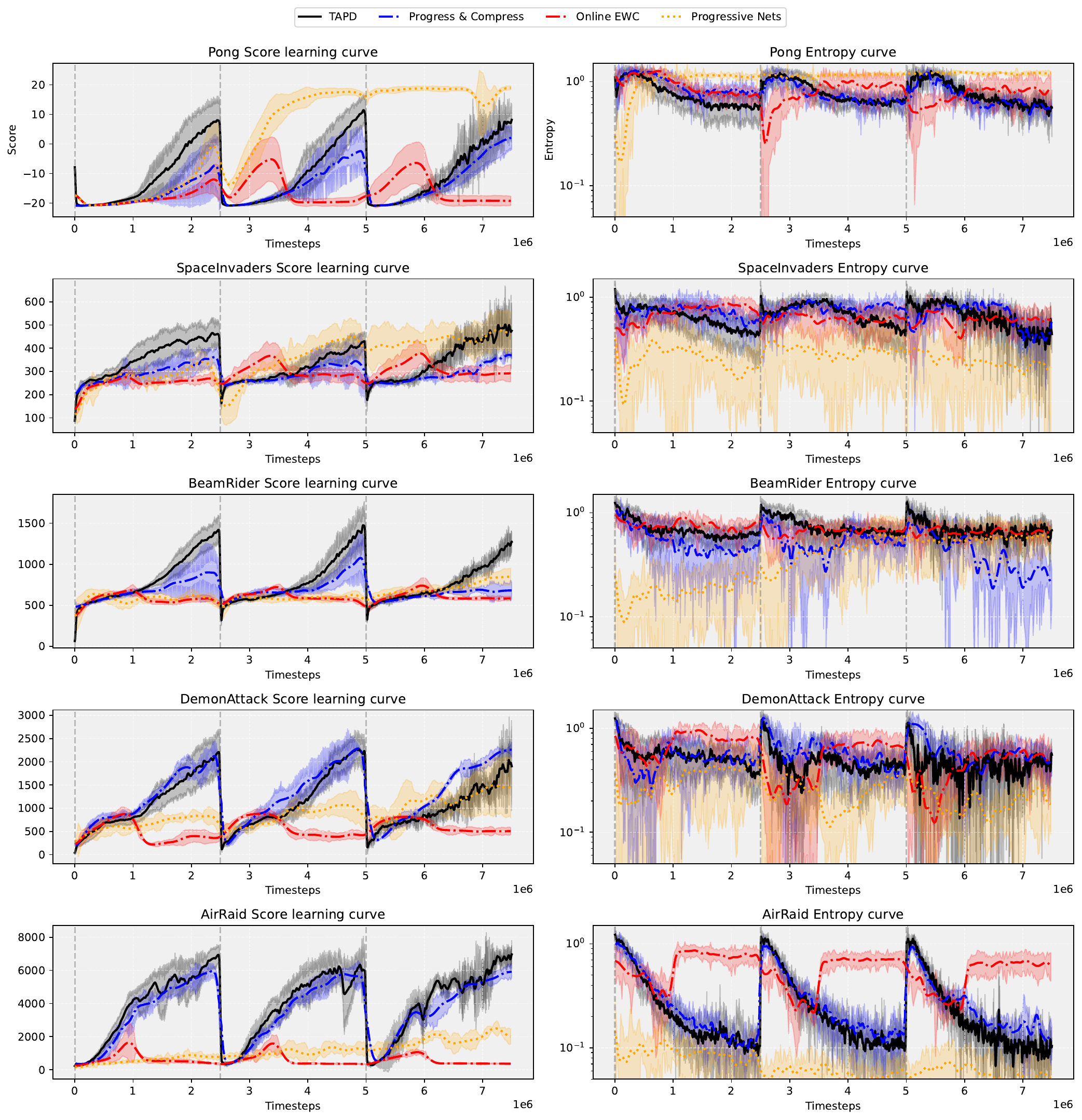}
        \end{subfigure}
        
    \end{minipage}
    
    \caption{The learning curves depicted represent the obtained rewards in the progress phase, against Task Agnostic Policy Distillation (TAPD), Online EWC, Progressive Nets, and the reproduced Progress \& Compress baseline. Reading from left to right, both performance and entropy are plotted. Tasks are learned in a sequential manner in the following order: Pong, SpaceInvaders, BeamRider, DemonAttack, and AirRaid. TAPD utilizes the distilled knowledge from the task-agnostic phase. Results are averaged over 4 seeds and reflect the averages of scores taken across 100 episodes. Each task is revisited three times (gray vertical lines), allowing for training for 2.5M timesteps on each visit in the progress phase.}
    \label{fig:baseline_comparison}
\end{figure}

Fig. \ref{fig:baseline_comparison} shows the learning curves of Pong, SpaceInvaders, BeamRider, DemonAttack, and AirRaid along with their corresponding entropies. In the case of Pong, although TAPD hadn't previously learned the game, its knowledge base enabled much faster exploration of newly encountered tasks, resulting in significantly higher performance than all other algorithms, which showed only slight improvement after 1.5 million timesteps. On the other hand, TAPD already learned to adapt to the tasks at around 0.6 million timesteps, showcasing the high sample efficiency of the proposed approach. This adaptation is also evident in the entropy of the distribution for solving the Pong task. The entropy initially starts lower than the Progress \& Compress baseline, indicating that the agent is aware of the required movements. However, the Progress \& Compress baseline exhibits a similar pattern throughout the training process, albeit with lower performance compared to TAPD. In later timesteps, TAPD exhibits a more exploratory behavior compared to the Progress \& Compress baseline, with occasional decreases. The knowledge base of TAPD seems to retain its exploratory behavior during the learning process of Pong. The entropy in Progressive Nets exhibits a steep learning curve during the second visit, which is expected. This is because the dedicated column used to train the Pong task in the first visit retains and continues to refine the parameters associated with Pong during the second visit, leading to rapid progress, which extends until the third and final visit. The Online EWC algorithm demonstrates a recurring pattern in the initial timesteps of each visit, where the score increases, indicating that the model detects the task but is unable to further improve upon it.

Moving on to the SpaceInvaders task, TAPD appears to remember this environment, which potentially contributes to its superior performance compared to all other algorithms in the initial phase. Throughout the training process, TAPD maintained lower entropy compared to the Progress \& Compress baseline, which showed much higher entropy across all three visits. Despite the Progress \& Compress baseline employing lateral connections in subsequent visits, its performance does not improve over time due to the high entropy of its policy, which suggests ongoing exploration. This does not mean the baseline is stuck, it may improve its performance after many more timesteps, but this would indicate lower sample efficiency. A similar behavior is observed with the Online EWC algorithm. The Progressive Nets algorithm demonstrates a strong initial performance with no interference from task-specific learned parameters, indicating a robust growth mechanism. However, despite the lack of interference in Progressive Nets, the TAPD algorithm ultimately converges to a higher performance by the end of the task visits. This suggests a positive forward transfer and a recurring pattern of stable entropy, contributing to its success.

In the BeamRider task, the TAPD algorithm consistently outperforms all other algorithms throughout the visits. Despite the Progressive Nets algorithm's use of dedicated task parameters, its performance and sample efficiency remain low, suggesting that lateral connections between columns and dedicated task parameters do not necessarily enhance forward transfer. In the AirRaid task, the Progress \& Compress baseline shows comparable performance to TAPD during the initial visits, but TAPD ultimately surpasses it by the end of the training. This suggests that the distillation of the intrinsically motivated behavior during the task-agnostic phase serves as a strong regularizer, enabling TAPD to achieve higher final performance. The smaller gap between visits in catching up to earlier performance suggests greater sample efficiency. In contrast, Progressive Nets algorithm struggles to catch up with the final performance of TAPD, further highlighting the advantages of the TAPD approach. In terms of entropy, it is evident that the TAPD algorithm begins with exploratory behavior that gradually decreases over the course of visits. This pattern suggests that the agent initially explores the environment more extensively compared to the more conservative action selection process of the Progressive Nets, which may contribute to its slower rate of performance improvement. Injecting task-agnostic behavior into the knowledge base during the task-agnostic phase appears to lead to a better trade-off between exploration and exploitation, enhancing overall performance and demonstrating that task-agnostic policy distillation indeed facilitates positive forward transfer.

Moreover, TAPD is significantly more scalable than Progressive Nets. TAPD requires only two networks, while Progressive Nets face a major limitation: as tasks are added, the network size increases substantially. Specifically, the number of hidden units and feature maps in Progressive Nets grows linearly with the number of columns, and the number of parameters grows quadratically. This scalability issue makes TAPD a more efficient and practical solution, especially when dealing with a large number of tasks. Additionally, Progressive Nets require the specific task ID during training to query the correct column in subsequent visits. This dependency on task identification is not necessary in TAPD, which makes it a more versatile method. TAPD can quickly adapt to tasks during the initial phases of training without needing explicit task IDs, further underscoring its flexibility and robustness compared to Progressive Nets.

\subsection{Assessing Forward Transfer}
\label{sec:fwd}

\begin{figure}
    \centering
    \includegraphics[width=0.7\linewidth]{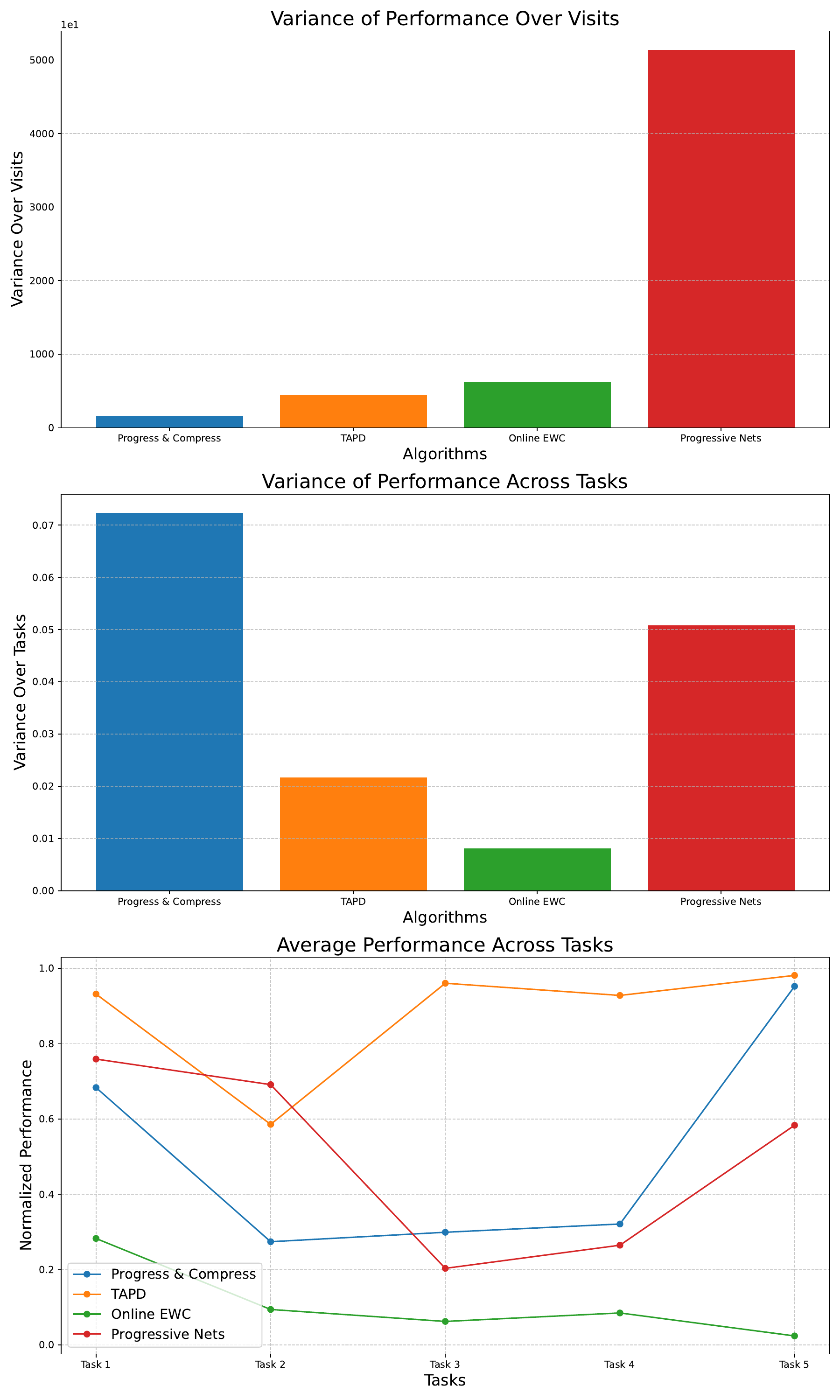}
    \caption{Analysis of Algorithm Performance: Assessing Forward Transfer through Variance Across Visits and Tasks and Average Performance Across Tasks. Averaged Over 8 Seeds.}
    \label{fig:fwd_transfer}
\end{figure}

Positive forward transfer refers to improved performance on a new task immediately following the learning of previous tasks, indicating an algorithm’s ability to leverage prior knowledge effectively. Two key indicators demonstrate positive forward transfer: (1) the performance of the active column across tasks over multiple visits, which captures improvement in learning a new task after exposure to prior tasks as well as gains from repeated visits to an old task, and (2) the average normalized performance of each algorithm across tasks, with high values suggesting successful application of previously acquired knowledge. Table \ref{tab:performance_variations} presents the performance data of the active column network across different tasks over three visits, comparing TAPD, Online EWC, Progressive Nets, and Progress \& Compress. This comparison evaluates how learning on one task influences subsequent task performance. In the table, an upward arrow ($\uparrow$) indicates improved performance compared to the previous visit, while a downward arrow ($\downarrow$) indicates decreased performance. Additionally, Fig. \ref{fig:fwd_transfer} illustrates the performance trends across multiple tasks and visits. Performance was evaluated using normalized scores, with further analysis focusing on the variance in performance across tasks and visits. As shown in Table \ref{tab:performance_variations}, TAPD outperforms all other algorithms in the three visits for the majority of tasks. In the third visit, it maintains the highest scores for Task 4 and Task 5, despite a slight decline in performance compared to the previous visit, as was also observed for Task 5 in the second visit.

\begin{table}[ht]
    \centering
    \caption{Performance of the active column network on subsequent tasks after visiting previous tasks. Tasks 1-5 were visited in the following order: Pong, SpaceInvaders, BeamRider, DemonAttack, and AirRaid, respectively. Results are averaged over 8 seeds.}
    \setlength{\tabcolsep}{4pt}
    \begin{tabular}{lccccc}
        \toprule
        & Task 1 & Task 2 & Task 3 & Task 4 & Task 5 \\
        \midrule
        \textbf{First Visit} \\
        \textit{Progress \& Compress (Active-Col)}      & -4.25 & 327.41 & 732.12 & 1029.89 & 2468.09\\
        \textit{TAPD (Active-Col)}                      & 14.86 & \textbf{439.62} & \textbf{1441.82} & \textbf{2479.65} & \textbf{2605.39} \\
        \textit{Online EWC}                             & 12.14 & 248.85 & 528.377 & 209.62 & 505.30 \\
        \textit{Progressive Nets}                             & \textbf{16.94} & 437.3 & 646.17 & 910.16 & 1279.05 \\
        \midrule
        \textbf{Second Visit} \\
        \textit{Progress \& Compress (Active-Col)}      & 14.56 $\uparrow$ & 335.97 $\uparrow$ & {877.66 $\uparrow$} &  {963.57 $\downarrow$} &  {2529.31 $\uparrow$} \\
        \textit{TAPD (Active-Col)}                      &  {\textbf{18.51} $\uparrow$} &  {\textbf{472.91} $\uparrow$} &  {\textbf{1537.52} $\uparrow$} &  {\textbf{2699.95} $\uparrow$} &  {\textbf{2580.15} $\downarrow$} \\
        \textit{Online EWC}                             &  {-14.71 $\downarrow$} &  {275.79 $\uparrow$} &  {606.64 $\uparrow$} &  {460.08 $\uparrow$} &  {433.57 $\downarrow$} \\
        \textit{Progressive Nets}                             &  {-3.4 $\downarrow$} &  {618.7 $\uparrow$} &  {697.6 $\uparrow$} & {612.37 $\downarrow$} &  {1455.93 $\uparrow$} \\
        \midrule
        \textbf{Third Visit} \\
        \textit{Progress \& Compress (Active-Col)}      &  {15.04 $\uparrow$} &  {387.19 $\uparrow$} &  {892.40 $\uparrow$} &  {1034.01 $\uparrow$} & {2501.98 $\downarrow$} \\
        \textit{TAPD (Active-Col)}                      &  {19.94 $\uparrow$} &  {\textbf{483.87} $\uparrow$} &  {\textbf{1550.30} $\uparrow$} &  {\textbf{2383.18} $\downarrow$} &  {\textbf{2506.41} $\downarrow$} \\
        \textit{Online EWC}                             &  {-17.19 $\downarrow$} &  {326.31 $\uparrow$} &  {641.08 $\uparrow$} &  {592.97 $\uparrow$} &  {390.29 $\downarrow$} \\
        \textit{Progressive Nets}                             &  {\textbf{20.33} $\uparrow$} &  {457.5 $\downarrow$} &  {864.68 $\uparrow$} &  {1084.24 $\uparrow$} &  {2312.06 $\uparrow$} \\
        \bottomrule
    \end{tabular}
    \label{tab:performance_variations}
\end{table}

The variance across visits (top plot in Fig. \ref{fig:fwd_transfer}) illustrates the variance of each algorithm's performance across multiple visits, providing insight into the stability of their learning processes. A high variance indicates that the algorithm's performance fluctuates significantly from one visit to the next, suggesting instability or sensitivity to specific conditions during each visit. Conversely, low variance indicates a more stable learning process, with the algorithm performing consistently and similarly across different visits. Progress \& Compress demonstrates the lowest variance, reflecting a stable performance across visits, making it potentially more reliable in scenarios requiring consistent outcomes. In contrast, the Progressive Nets algorithm shows the highest variance over visits, indicating significant performance fluctuations and suggesting overfitting during each task visit.

The variance across tasks (middle plot in Fig. \ref{fig:fwd_transfer}) measures how consistently each algorithm performs across different tasks. Progress \& Compress and Progressive Nets exhibit higher variance, indicating that their performance is uneven across tasks. This suggests that these algorithms may excel in certain types of tasks but struggle with others, leading to a less predictable overall performance. High variance is a sign of overfitting to specific task characteristics, limiting the algorithms generalization capabilities across diverse tasks. This outcome is expected, as the model's complexity increases with each new task and training iteration in Progressive Nets. In contrast, low variance suggests that the algorithm performs more uniformly across different tasks, which is desirable for generalization. TAPD, with its lower variance, demonstrates a more balanced and consistent performance across different tasks, indicating better generalization.

The bottom plot in Fig. \ref{fig:fwd_transfer} provides an overview of the average normalized performance of each algorithm across different tasks, serving as a measure of forward transfer. This comparison highlights how well each algorithm performs relative to the others on the same task. Forward transfer represents an algorithm's capability to leverage knowledge from previous tasks and visits to improve performance on new ones. An upward or stable trend in this plot suggests successful application of previously acquired knowledge to subsequent tasks, demonstrating positive forward transfer.

All algorithms demonstrate relatively stable and consistent performance across tasks, indicating effective forward transfer. Notably, Online EWC, despite its lower overall performance, shows exceptional stability and consistency, suggesting strong generalization capabilities. In contrast, Progressive Nets exhibit greater fluctuation, indicating that their forward transfer through lateral connections is less effective, leading to variable outcomes depending on the task. As the network expands, the features of new columns tend to become less significant overall \cite{rusu2016progressive, xu2021adaptive}. TAPD consistently outperforms all other algorithms, demonstrating superior and faster task adaptability. This is evidenced by its consistently stable performance curve, which also highlights TAPD's greater sample efficiency.

The variability in the performance of Progressive Nets may stem from model complexity and the use of new lateral connections, raising concerns about its ability to consistently generalize across tasks. Online EWC distinguishes itself with stability over time and consistent performance, indicating strong forward transfer and generalization capabilities. Both Progress \& Compress and TAPD offer balanced performance across tasks, with TAPD being particularly reliable for diverse tasks due to its faster adaptability. An algorithm like TAPD, which maintains low variance across tasks and visits, is demonstrably more versatile and effective at handling a variety of challenges, making it especially well-suited for environments with diverse task demands.

\subsection{Computational Efficiency}
TAPD is a multi-phase process designed to balance computational demands with performance gains. During the initial task-agnostic phase, TAPD interacts with the environment for approximately 300,000 timesteps per task (game environment), with the total number of timesteps being dependent on the number of environments (denoted by $x$). For example, when $x = 25$, this corresponds to a total of 7.5 million timesteps, distributed uniformly across all environments. This phase, while computationally intensive and requiring significant interaction with the environment, is manageable within the overall process. Specifically, processing 7.5 million timesteps can take approximately 80 minutes, as shown in Table \ref{tab:comp_time}. Although this phase is time-consuming and represents the computational bottleneck in TAPD, it lays a crucial foundation for subsequent performance improvements. It is important to note that, by design, TAPD can operate without externally specified tasks, distinguishing it from other methods that lack this task-agnostic capability.

After completing the data collection from the task-agnostic phase, TAPD compresses the policy for each task sample. This compression step, while less computationally demanding, is critical for ensuring that the learned policy generalizes effectively across tasks. In this step, TAPD computes the Fisher information matrix for the compressed policy. Although the computation involves around 100 updates with a minibatch size of 32 \cite{kirkpatrick2017overcoming} and is relatively low in computational cost, the accuracy of the Fisher matrix might not be perfect but remains sufficient for the intended purpose.

Progressive Nets generally require a high computational cost due to their architecture. Each new task involves adding new network components, which increases the computational burden and memory requirements as more tasks are added. While they may offer strong performance on individual tasks during training, Progressive Nets are severely limited by their inability to scale, with model size growing excessively with each new task. The variance in performance in test time over visits indicates instability, as discussed in Section \ref{sec:fwd}, which might require additional computational resources to mitigate.

Online EWC is less computationally demanding compared to Progressive Nets. It leverages the Fisher information matrix to regularize the network weights, preventing catastrophic forgetting. The computational cost primarily arises from calculating the Fisher matrix, similar to TAPD but with a simpler model.

Despite the high initial computational cost in the task-agnostic phase, TAPD provides a strong balance between computational efficiency and performance. The time invested in this phase is offset by the improved forward transfer and generalization capabilities observed in subsequent tasks. This makes TAPD an excellent choice when the goal is to maximize performance after an initial warm-up period.

\begin{table}[h!]
\centering
\resizebox{\textwidth}{!}{%
\begin{tabular}{|l|p{8cm}|c|}
\hline
\textbf{Algorithm} & \textbf{Training phases} & \textbf{Time} \\ \hline

\multirow{3}{*}{\textbf{TAPD}} & 
\begin{tabular}[c]{@{}l@{}} 
Progress: DemonAttack, Pong, BeamRider, SpaceInvaders, \\AirRaid (2.5M timesteps)
+ Compress (150,000 timesteps)
\end{tabular} & $\sim$ 20 minutes (each visit) \\ 
& Task-Agnostic Phase & $\sim$ 80 minutes \\
& \textbf{Total Time} & $\sim$ \textbf{6.3 hours} \\ \hline

\multirow{2}{*}{\textbf{Progress \& Compress}} & 
\begin{tabular}[c]{@{}l@{}} 
Progress: DemonAttack, Pong, BeamRider, SpaceInvaders, \\AirRaid (2.5M timesteps)
+ Compress (150,000 timesteps)
\end{tabular} & $\sim$ 20 minutes (each visit) \\ 
& \textbf{Total Time} & $\sim$ \textbf{5.3 hours} \\ \hline

\textbf{Progressive Nets} & \textbf{Total Time} & $\sim$ \textbf{5.3 hours} \\ \hline
\textbf{Online EWC} & \textbf{Total Time} & $\sim$ \textbf{4.3 hours} \\ \hline

\multicolumn{3}{|c|}{\textbf{Hardware:} NVIDIA GeForce GTX 1660 SUPER and Intel Core i5-10500 CPU @ 3.10GHz} \\ \hline
\end{tabular}
}
\caption{Computational Time Comparison of Various Algorithms on NVIDIA GeForce GTX 1660 SUPER and Intel Core i5-10500 CPU. The total time includes training and other phases without any intermediate evaluations.}
\label{tab:comp_time}
\end{table}

\section{Conclusion and Future Work}
In this paper, we presented the Task-Agnostic Policy Distillation framework, which addresses catastrophic forgetting, ensures scalability across tasks, enables positive forward transfer, and facilitates learning without requiring task labels. The framework incorporates a task-agnostic phase within the algorithmic framework of Progress and Compress proposed by \citet{schwarz2018progress}. This task-agnostic phase uses a self-supervised prediction error as an intrinsic reward for the agent. By doing so, the agent learns without a specific reward function and does not require clear task boundaries. The task-agnostic phase can be implemented in different variations that abstract the task-agnostic phase as a process where task-agnostic policies are distilled into the knowledge base, increasing systematic exploratory behavior. The active column in the task-agnostic phase then utilizes this knowledge base, further maximizing its intrinsic reward based on the systematic exploratory behavior. This acts as a pre-training phase before downstream tasks are applied. 

It has been shown that the most general variation of the task-agnostic phase improves performance by accelerating transfer learning in the progress phase for downstream tasks, surpassing all three continual learning baselines, including the Progress and Compress method, in the Atari domain. Consequently, the Task-Agnostic Policy Distillation framework has demonstrated promising results in enhancing positive forward transfer and learning in scenarios without clear task boundaries.

Our approach addresses catastrophic forgetting by using Elastic Weight Consolidation (EWC) to protect old knowledge. Positive forward transfer is achieved through lateral connections from the knowledge base to the active column. Scalability across tasks is ensured by utilizing a single policy network for retaining old knowledge of previous tasks. This makes our framework well-suited to environments with restricted memory and onboard resources. Learning without clear task boundaries is facilitated by a task-agnostic phase that encourages exploration without relying on task-specific extrinsic rewards.

While the obtained results are promising, there is potential for further improvement. This includes exploring new variations of the task-agnostic phase. It would also be valuable to analyze variations within other domains. For instance, applying the general Variation 1 to domains such as robotic tasks. Furthermore, incorporating the intrinsic reward during the progress phase on a downstream task allows for optimizing both extrinsic and intrinsic rewards in an alternating fashion. This approach would likely be especially advantageous for long-horizon robotic tasks, where balancing exploration and exploitation is challenging, as rewards for exploration may not be received immediately.

\section*{Declarations}
\textbf{Conflicts of interest.} The authors have no relevant financial or non-financial interests to disclose.
\newline
\noindent\textbf{Data availability statement.} The datasets used and/or analysed during the current study available from the corresponding author on reasonable request.
\newline
\noindent\textbf{Code availability statement.} The implementation of our TAPD algorithm, along with the baseline methods and the code to reproduce our experiments, is available at the following repository: \url{https://github.com/wabbajack1/TAPD}.


\begin{appendices}

\section{Progress and Compress Phases}\label{secA1}
\textbf{Progress.} The progress phase is designed to effectively minimize the loss of a specified task, i.e.,  \(\mathcal{L}^{progress} = -\mathbb{E}_\pi \left[ \sum_{n=0}^{\infty} \gamma^n r_{t+n} \right]\), which is equivalent to the RL objective \(\max_{\pi} \mathbb{E}_\pi \left[ \sum_{n=0}^{\infty} \gamma^n r_{t+n} \right]\), where $r$ is the reward and $\gamma$ is the discount factor. As in \citet{schwarz2018progress}, we used a distributed variant of the actor-critic architecture (A2C) to learn both the policy \(\pi(a_t|s_t; \theta_1)\) and the value function \(V(s_t; \theta_2)\), from raw pixels, within both phases. The policy and value function (both referred to as \textit{active column}) share a convolutional encoder, similar to the approach described by \citet{mnih2016asynchronous}. 

\noindent\textbf{Compress.} The compress phase incorporates a variety of methods, making the implementation and fine-tuning of hyperparameters the most time-intensive aspect of this process. Unlike the progress phase, during which the primary focus is on one task, the compress phase is crucial for retaining knowledge as tasks are encountered sequentially. After acquiring knowledge for a designated number of timesteps during the progress phase, the most recent policy (the active column) of the current task is distilled into the knowledge base network (the same copy of the network architecture was used as in the progress phase) by minimizing the following distillation loss with respect to the parameters $\theta_{kb}$ of the knowledge base:

\begin{equation} \label{eq:10}
\mathcal{L}^{Distill}(\theta_{kb}) = \mathbb{E}\left[ D_{\text{KL}}\left( \pi_k(\cdot | x; \theta_{active}) \parallel \pi_{\textit{kb}}(\cdot | x; \theta_{kb}) \right) \right] + \frac{\lambda}{2} \gamma F_{k-1} \left\|\theta_{kb} - \theta_{kb}^{*(k-1)} \right\|_2^2
\end{equation}

\noindent where \(x\) is the input and \( \pi_k(\cdot | x; \theta_{active}) \), \( \pi_{\textit{kb}}(\cdot | x; \theta_{kb}) \) are the policies of the active column (after learning on task \( k \)) and the knowledge base, respectively. \(F_{k-1}\) is the diagonal Fisher Information Matrix (FIM) of the previous tasks and \(\gamma > 0\) is the forgetting constant, introducing gracefully forgetting old knowledge (mind the overloaded notation $\gamma$). \(\lambda\) is the importance of the penalty term. \(\theta_{kb}^{*(k-1)}\) are the optimal parameters of the previous tasks. The loss function \(\mathcal{L}^{Distill}(\theta_{kb})\) is then estimated by stochastic gradient descent, finding \(\theta^{*}_{kb} = \arg\min_{\theta_{kb}} \mathcal{L}^{Distill}(\theta_{kb})\). This optimization is referred to in \citet{schwarz2018progress} as online Elastic Weight Consolidation or Online EWC in short. It is important that the penalty can only be applied after the first compress phase. The knowledge base not only acts as a repository of cumulative knowledge but also enables the active column to build upon this consolidated information. The integrated knowledge serves as a foundation for the model to further refine and optimize its learning processes. This iterative learning allows for enhancing the model's performance on subsequent tasks.

Before any subsequent compress phase starts, the retainment of the knowledge base's current parameters \(\theta^*_{kb}\) is necessary. For doing this, it is important that after the very first compress phase, the diagonal of the Fisher information matrix gets estimated with respect to the latest compressed optimal parameters \(\theta^*_{kb}\) of the knowledge base network. As the A2C is an on-policy algorithm, in order to calculate the Fisher estimate w.r.t. \(\theta^*_{kb}\), we need to generate samples based on the policy \(\pi_{kb}(\cdot|x;\theta^*_{kb})\) to have an accurate estimation of the diagonal of the Fisher information. Estimating the Fisher information is as follows:
\begin{equation}
\label{eq:estimatefisher}
    F(\theta^*_{kb}) = \frac{1}{N} \sum_{t=1}^{N} \left( \frac{\partial (\log(\pi_{kb}(a_t|s_t;\theta^*_{kb})) \cdot A_t)}{\partial \theta^*_{kb}} \right)^2
\end{equation}
where \(N\) is the number of samples generated by the policy \(\pi_{kb}(\cdot|x;\theta^*_{kb})\). The algorithmic procedure is outlined in Algorithm \ref{alg:dist_online}.

\begin{algorithm}
\caption{Distillation + Online EWC Algorithm}
\label{alg:dist_online}
\begin{algorithmic}[1]
\State \textbf{Input:} Tasks $\{1, 2, \ldots, T\}$, forgetting factor $\gamma$, regularization hyperparameter $\lambda$
\State \textbf{Initialize:} Distill (train) Task 1 to obtain $\theta^{*(1)}_{kb} = \arg\min_{\theta_{kb}} \mathcal{L}^{Distill}_1(\theta_{kb})$
\For{each task $t > 1$}
    \State \textbf{Compute FIM:} $F_{t-1} = \mathbb{E}_{s_{t-1} \sim P_{t-1}(\cdot)}\left[ \left( \frac{\partial (\log(\pi_{kb}(a_{t-1}|s_{t-1};\theta_{kb}^{*(t-1)})) \cdot A_{t-1})}{\partial \theta_{kb}^{*(t-1)} }\right)^2 \right]$
    \State \textbf{Online FIM Update:} $F_{t-1} = \gamma F_{t-2} + F_{t-1}$
    \State \textbf{Distillation Loss:} $\mathcal{L}_{t}^{Distill}(\theta_{kb}) = \mathcal{L}_{t}(\theta_{kb}) + \frac{\lambda }{2} F_{t-1} \left\|\theta_{kb} - \theta_{kb}^{*(t-1)} \right\|_2^2$
    \State \textbf{Update Parameters:} $\theta_{kb}^{*(t)} = \arg\min_{\theta_{kb}} \mathcal{L}_{t}^{Distill}(\theta_{kb})$
\EndFor
\end{algorithmic}
\end{algorithm}

\noindent\textbf{Architecture.}
An identical architecture is used within both the progress and compress phases. The active column consists of one policy \(\pi_{\textit{active}}(a_t|s_t; \theta^1_{\textit{active}})\) and a value function \(V_{\textit{active}}(s_t; \theta^2_{\textit{active}})\) which share a convolutional encoder \({\phi_\textit{active}}(\cdot; \theta_{active}^3)\). This encoder is represented by a convolutional neural network (CNN), following the approach outlined by \citet{mnih2016asynchronous}. Let \( s_t \) represent the sequence of observations at timestep \( t \) from the environment, where \(\phi_{\textit{active}}(s_t)\) indicates the encoded visual features used for subsequent policy learning. The same principle applies to the knowledge base network, with the only difference being the change in the subscript to \(\textit{kb}\). It is important to note that the active column incorporates lateral connections from the knowledge base. This is done through the function \(f(\cdot)\) represented by a neural network consisting of convolutional layers and a multi-layer perceptron with one hidden layer and designed to process the intermediate outputs from the knowledge base. These intermediate outputs are derived from \(\phi_{\textit{kb}}(s_t)\), \(\pi_{\textit{kb}}(a_t|s_t; \theta^1_{\textit{kb}})\), and \(V_{\textit{kb}}(s_t; \theta^2_{\textit{kb}})\). Initially, during the training of the first task for a specified number of timesteps, the active column does not utilize any lateral connections, keeping the knowledge base inactive. However, after the compress phase, the lateral connections are activated, utilizing prior knowledge to optimize both the active column and the lateral connections in the progress phase of the next task.

\noindent\textbf{Learning in the Progress and Compress Framework.}
Algorithm \ref{alg:pc} represents the approach of \citet{schwarz2018progress} in the RL domain, unifying all arguments mentioned in the previous sections. In every iteration, a vectorized environment is constructed corresponding to each task.

\begin{algorithm}[H]
\caption{Vanilla Progress and Compress Algorithm}
\label{alg:pc}
\begin{algorithmic}[1]
\State \textbf{Initialize:} policy networks $\pi_{active}$, $\pi_{kb}$; encoders $\phi_{active}$, $\phi_{kb}$; tasks $\mathcal{T}$; timesteps $T_{active}, T_{kb}, T_{Fisher}$; buffer $U$; encoder $f$; visits $\mathcal{V}$
\For{each visit $v$ in $\mathcal{V}$}
    \For{each task $k$ in $\mathcal{T}$}
        \State Set up environment $E^k$
        \State Unfreeze $\pi_{active}, \phi_{active}, f$; Freeze $\pi_{kb}, \phi_{kb}$
        \For{$t = 1$ to $T_{active}$}  \Comment{Progress Phase}
            \State Observe $s_t \sim E^k$, $\phi = \phi_{active}(s_t)$, $a_t \sim \pi_{active}(f(\phi))$
            \State $r_{t+1}, s_{t+1} \sim P^{E^k}(s_t, a_t)$, $U \gets U \cup \{(r_{t+1}, s_t, s_{t+1}, a_t)\}$
            \If{$t \mod \text{size}(U) = 0$}
                \State Evaluate actions based on rollout $U$
                \State Update $\theta_{active}$ using SGD; Clear $U$ \Comment{the loss $\mathcal{L}^{progress}(\theta_{active})$}
            \EndIf
        \EndFor
        \State Switch modes: Freeze $\pi_{active}, \phi_{active}$; Unfreeze $\pi_{kb}, \phi_{kb}$
        \For{$t = 1$ to $T_{kb}$} \Comment{Compress Phase}
            \State Observe $s_t \sim E^k$, $\phi = \phi_{kb}(s_t)$, $a_t \sim \pi_{kb}(\phi)$
            \State $s_{t+1} \sim P^{E^k}(s_t, a_t)$, $U \gets U \cup \{(s_t, s_{t+1}, a_t)\}$
            \If{$t \mod \text{size}(U) = 0$}
                \If{$k <$ 2}
                    \State Update $\theta_{kb}$ with SGD on {\small $D_{\text{KL}}(\pi_{active}(\cdot|s_{t:\text{size}(U)})\parallel \pi_{kb}(\cdot|s_{t:\text{size}(U)}))$}
                \Else
                    \State Update $\theta_{kb}$ with SGD on {\small $D_{\text{KL}}(\pi_{active}(\cdot|s_{t:\text{size}(U)})\parallel \pi_{kb}(\cdot|s_{t:\text{size}(U)}))$}
                   \newline {\small $+ \frac{\lambda}{2} F_{k-1} \left\|\theta_{kb} - \theta_{kb}^{*(k-1)}\right\|_2^2$}
                \EndIf
                \State Clear $U$
            \EndIf
        \EndFor
        \For{$t = 1$ to $T_{Fisher}$}
            \State Observe $s_t \sim E^k$, $\phi = \phi_{kb}(s_t)$, $a_t \sim \pi_{kb}(\phi), s_{t+1} \sim P^{E^k}(s_t, a_t)$
            \State $U \gets U \cup \{(s_t, s_{t+1}, a_t)\}$
            \If{$t \mod \text{size}(U) = 0$}
                \State Estimate Fisher information based on $U$; Clear $U$
            \EndIf
        \EndFor
        \State Update Fisher information for $\theta^*_{kb}$; Reinitialize $\theta_{active}$ \Comment{see Equation \ref{eq:estimatefisher}}
    \EndFor
\EndFor
\end{algorithmic}
\end{algorithm}




\section{Architecture Details}\label{secA2}%

\subsection{A2C Hyperparameters}
The following hyperparameters (see Table \ref{tab:a2c_architecture}) specify both the active column and knowledge base networks, which are used in both progress and compress phases for representing the actor (policy) and critic (value function).

\begin{table}[h]
\centering
\caption{Neural network architecture of the active column and knowledge base}
\label{tab:a2c_architecture}
\begin{tabular}{|c|l|l|l|l|l|}
\hline
\textbf{Order} & \textbf{Layer Type} & \textbf{Activation} & \textbf{Size} & \textbf{Filter Size} & \textbf{Filter Stride} \\
\hline
1 & Convolution & ReLU & 32 & [8 × 8] & [4 × 4] \\
\hline
2 & Convolution & ReLU & 64 & [4 × 4] & [2 × 2] \\
\hline
3 & Convolution & ReLU & 32 & [3 × 3] & [1 × 1] \\
\hline
4 & Flatten & - & - & - & - \\
\hline
5 & Fully-Connected & ReLU & \text{512} Neurons & - & - \\
\hline
6 & Critic & Linear & 1 Neuron & - & - \\
\hline
7 & Actor(Policy) & Linear & 4 Neurons & - & - \\
\hline
\end{tabular}
\end{table}

\subsection{Forward Model Hyperparameters}
The following hyperparameters (see Table \ref{tab:nn_architecture}) are utilized to assemble the forward model incorporated in the implementation. In this process, the action is one-hot encoded and subsequently merged with an encoded state feature representation.

\begin{table}[h]
\centering
\caption{Neural network architecture of the forward model}
\label{tab:nn_architecture}
\begin{tabular}{|c|l|l|l|l|l|}
\hline
\textbf{Order} & \textbf{Layer Type} & \textbf{Activation} & \textbf{Size} & \textbf{Filter Size} & \textbf{Filter Stride} \\
\hline
1 & Convolution & ReLU & 32 & [3 × 3] & [2 × 2] \\
\hline
2 & Convolution & ReLU & 32 & [3 × 3] & [2 × 2] \\
\hline
3 & Convolution & ReLU & 32 & [3 × 3] & [2 × 2] \\
\hline
4 & Convolution & ReLU & 32 & [3 × 3] & [2 × 2] \\
\hline
5 & Convolution & ReLU & 32 & [3 × 3] & [2 × 2] \\
\hline
6 & Fully-Connected & ReLU & 256 Neurons & - & - \\
\hline
7 & Fully-Connected & Linear & 288 Neurons & - & - \\
\hline
\end{tabular}
\end{table}

\subsection{Hyperparameters of the Experiments}
All variations in timesteps across experiments are explicitly stated. One important parameter is the ``num-samples-drawn-in-task-agnostic-phase" (see Table \ref{tab:parameter_comparison}). This represents a procedure in the task-agnostic phase where games are randomly selected and trained for a specific duration set by ``num-env-steps-agnostic". Then, the compress phase is executed for ``num-env-steps-compress-agnostic". This process is repeated 30 times, where 30 samples are uniformly drawn from the Meta-Environment. The Meta-Environment includes Pong (P), SpaceInvaders (S), and BeamRider (B). The RMSprop optimizer was consistently used in all experiments.

\begin{table}
    \label{hyper}
    \centering
    \small
    \renewcommand{\arraystretch}{1.2}
    \setlength{\tabcolsep}{4pt} 
    \caption{Hyperparameter values in the Progress \& Compress baseline, TAPD, Progressive Nets, and Online EWC of the task-agnostic phase}
    \begin{tabular}{|l|c|c|c|c|}
        \hline
        \textbf{Parameter} &  \textbf{Baseline} & \makecell{\textbf{Task-Agnostic} \\ \textbf{Policy Distillation} \\ \textbf{(TAPD)}} & \makecell{\textbf{Progressive} \\ \textbf{Nets}} & \makecell{\textbf{Online} \\ \textbf{EWC}} \\
        \hline
        agnostic-phase & - & True & - & - \\
        \hline
        batch-size-fisher (Fisher information estimation) & \multicolumn{4}{c|}{32} \\
        \hline
        eval-steps & \multicolumn{4}{c|}{$10^5$} \\
        \hline
        ewc-lambda & \multicolumn{4}{c|}{2} \\
        \hline
        ewc-gamma & \multicolumn{4}{c|}{0.3} \\
        \hline
        gamma (Discount factor for rewards) & \multicolumn{4}{c|}{0.99} \\
        \hline
        ewc-start & \multicolumn{4}{c|}{$15 \times 10^4$} \\
        \hline
        entropy-coef (Entropy term coefficient) & \multicolumn{4}{c|}{0.01} \\
        \hline
        lr (Learning rate) & \multicolumn{4}{c|}{$7 \times 10^{-4}$} \\
        \hline
        eps (RMSprop optimizer epsilon) & \multicolumn{4}{c|}{$1 \times 10^{-5}$} \\
        \hline
        alpha (RMSprop optimizer alpha) & \multicolumn{4}{c|}{0.99} \\
        \hline
        num-env-steps-agnostic & - & $3 \times 10^5$ & - & - \\
        \hline
        num-env-steps-compress ($T_{kb}$) & \multicolumn{2}{c|}{$3 \times 10^5$} & - & - \\
        \hline
        num-env-steps-agnostic-compress & - & $3 \times 10^5$ & - & -\\
        \hline
        num-env-steps-progress ($T_{active}$) & \multicolumn{4}{c|}{$2.5 \times 10^6$} \\
        \hline
        num-processes & \multicolumn{4}{c|}{10} \\
        \hline
        num-visits & \multicolumn{4}{c|}{3} \\
        \hline
        value-loss-coef & \multicolumn{4}{c|}{0.5} \\
        \hline
        max-grad-norm & \multicolumn{4}{c|}{0.5} \\
        \hline
        num-steps-fisher ($T_{Fisher}$) [11] & \multicolumn{4}{c|}{100} \\
        \hline
        num-steps (Rollout size) & \multicolumn{4}{c|}{20} \\
        \hline
        num-samples-drawn-in-task-agnostic-phase & - & 25 & - & - \\
        \hline
        Tasks in task-agnostic phase & - & S, B & - & - \\
        \hline
        \makecell{\text{Tasks in progress and compress phases} \\ / \text{Training phase} } & \multicolumn{4}{c|}{\makecell{Pong (P), SpaceInvaders (S), BeamRider (B), \\ DemonAttack (D), AirRaid (A)}} \\
        \hline
    \end{tabular}
    \label{tab:parameter_comparison}
\end{table}

\end{appendices}

\bibliography{sn-bibliography}

\end{document}